\documentclass[10pt,twocolumn,letterpaper]{article}

\usepackage[pagenumbers]{wacv} 

\usepackage{graphicx}
\usepackage{amsmath}
\usepackage{amssymb}
\usepackage{booktabs}

\usepackage{times}
\usepackage{epsfig}
\usepackage{multirow}
\usepackage{diagbox}
\usepackage{xcolor}

\usepackage[pagebackref,breaklinks,colorlinks]{hyperref}

\usepackage[capitalize]{cleveref}
\crefname{section}{Sec.}{Secs.}
\Crefname{section}{Section}{Sections}
\Crefname{table}{Table}{Tables}
\crefname{table}{Tab.}{Tabs.}

\def\wacvPaperID{2300} 
\def\confName{WACV}
\def\confYear{2024}

\usepackage[accsupp]{axessibility} 

\begin{document}

\title{Hyb-NeRF: A Multiresolution Hybrid Encoding for Neural Radiance Fields}

\author{Yifan Wang\\
 SUSTech\\
{\tt\small wangyf1998sc@foxmail.com}
\and
 Yi Gong \footnotemark[2] \\
 SUSTech\\
 {\tt\small gongy@sustech.edu.cn}
\and
 Yuan Zeng \footnotemark[2] \\
 SUSTech\\
 {\tt\small zengy3@sustech.edu.cn}
}

\maketitle

\footnotetext{$\dagger$ Corresponding authors}

\begin{abstract}
Recent advances in Neural radiance fields (NeRF) have enabled high-fidelity scene reconstruction for novel view synthesis. However, NeRF requires hundreds of network evaluations per pixel to approximate a volume rendering integral, making it slow to train. Caching NeRFs into explicit data structures can effectively enhance rendering speed but at the cost of higher memory usage. To address these issues, we present Hyb-NeRF, a novel neural radiance field with a multi-resolution hybrid encoding that achieves efficient neural modeling and fast rendering, which also allows for high-quality novel view synthesis. The key idea of Hyb-NeRF is to represent the scene using different encoding strategies from coarse-to-fine resolution levels. Hyb-NeRF exploits memory-efficiency learnable positional features at coarse resolutions and the fast optimization speed and local details of hash-based feature grids at fine resolutions. In addition, to further boost performance, we embed cone tracing-based features in our learnable positional encoding that eliminates encoding ambiguity and reduces aliasing artifacts. Extensive experiments on both synthetic and real-world datasets show that Hyb-NeRF achieves faster rendering speed with better rending quality and even a lower memory footprint in comparison to previous state-of-the-art methods. 
\end{abstract}

\section{Introduction}

\begin{figure}[h]
\begin{center}
\includegraphics[width=1.00\linewidth]{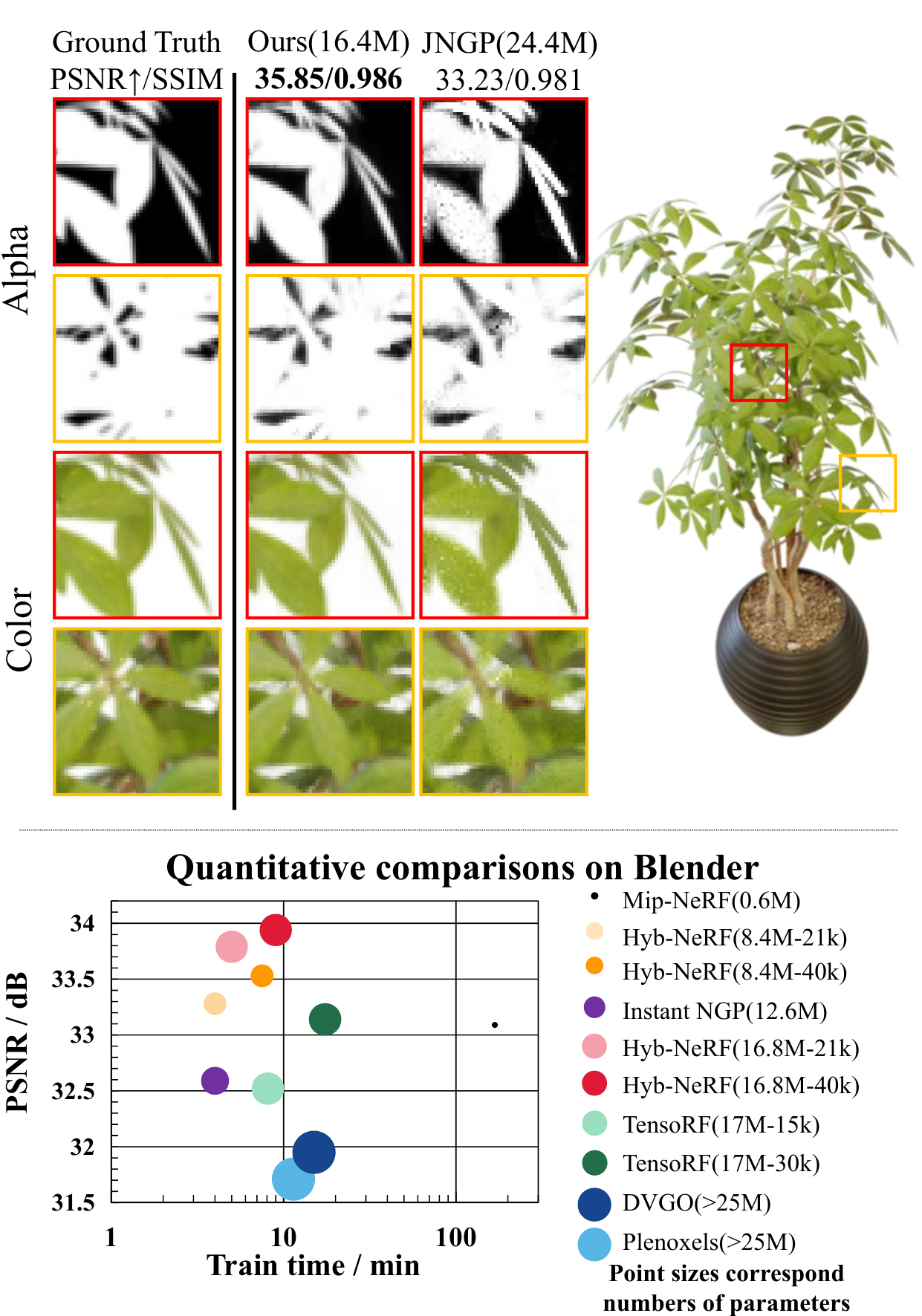}
\end{center}
\caption{Top: Rendering performance comparison between our Hyb-NeRF and JNGP\ \cite{yang2023jnerf} on the \emph{ficus} scene in Blender. Our Hyb-NeRF can render high-quality color images and synthesize an alpha map with less noise, enabling to accurate reconstruction of the translucency of the leaf edges without jaggies. Bottom: In comparison with previous state-of-the-art fast NeRF training methods, our Hyb-NeRF models can achieve the best rendering quality and memory compactness, while maintaining fast rendering.
}
\label{fig:frist_img}
\end{figure}

\begin{figure}[h]
\begin{center}
\includegraphics[width=1\linewidth]{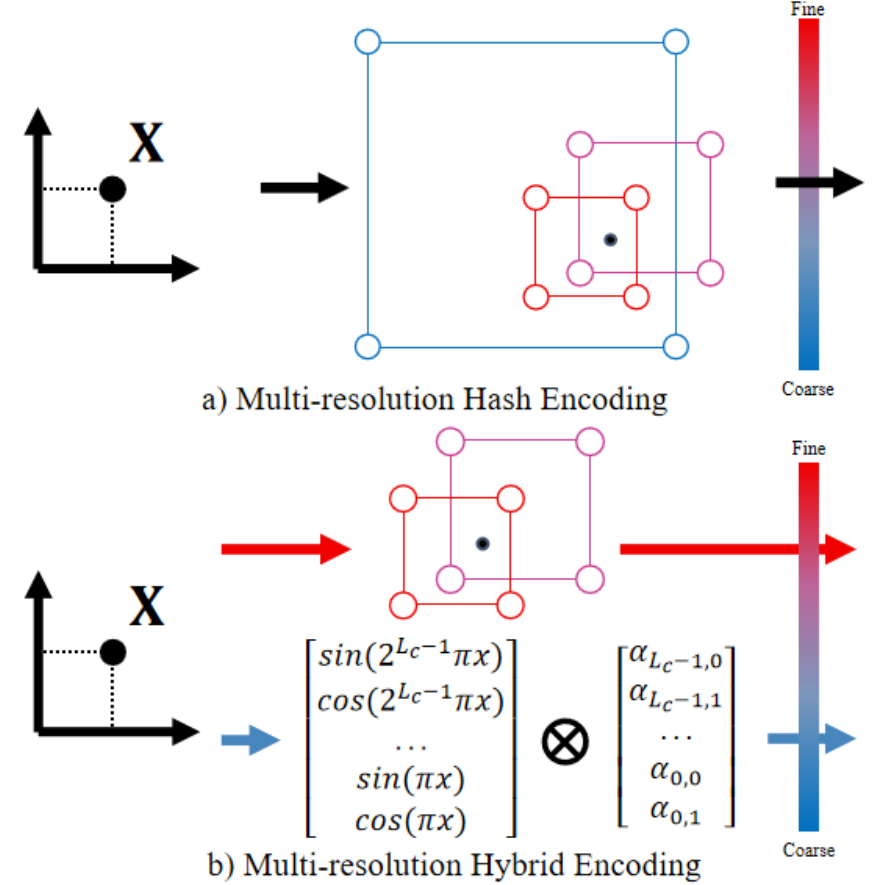}
\end{center}
\caption{a) Multi-resolution hash encoding\ \cite{muller2022instant} maps the input position $\mathbf{x}$ to hash-based feature grids at all resolution levels. b) Our multi-resolution hybrid encoding uses the concatenation of coarse-level learnable positional features and fine-level hash-based dense grids of trainable features to represent the position $\mathbf{x}$, resulting in a significantly lower memory footprint and higher quality representation of synthetic and real-world scenes.
}
\label{fig:fristimg2}
\end{figure}

Novel view synthesis targets rendering a scene from a set of images and camera poses obtained from unobserved viewpoints. Synthesizing novel views in real-time at photo-realistic quality is a long-standing problem in computer vision and computer graphics. To address this problem, traditional approaches like rasterization and ray-tracing rely on feature matching and view interpolation, requiring significant manual effort in designing and pre-processing the scene. Recently, Neural Radiance Fields (NeRF) and its variants\ \cite{mildenhall2021nerf,barron2021mip,chen2022tensorf,liu2020neural} have shown impressive performance on scene representation and novel view synthesis. These approaches obtain high-quality rendering of scenes by implicitly encoding colors and volume densities using coordinate-based multi-layer perceptrons (MLPs).

A key commonality of NeRF-like models is to encode the low-dimensional coordinate to a higher-dimensional space that assists the MLPs in learning scene representation more accurately. NeRF and its variants\ \cite{mildenhall2021nerf, barron2021mip} encode the input 5D coordinate as a multi-resolution sequence of Fourier features using fixed positional encodings, allowing the MLPs to capture high-frequency details that are essential for photo-realistic novel view synthesis. Despite the significant progress in representing high-frequency details of scenes, NeRF and its variants still require a large MLP to transform the non-parametric features into color and density, which requires a lengthy time for training and rendering. 


To address the computational efficiency issue, caching neural radiance fields into explicit data structures has been considered in recent works\ \cite{yu2021plenoctrees, fridovich2022plenoxels, reiser2023merf, chen2022tensorf, sun2022direct}. By caching additional trainable parameters in an auxiliary data structure, these approaches can learn color and density with a much shallower MLP, which improves rendering speed at the cost of storing a large set of features in a discrete data structure. Previous methods\ \cite{yu2021plenoctrees,fridovich2022plenoxels} have shown that high-resolution grids with resolution initialization and progressive interpolation work well on scene representation. Instead of doing progressive interpolation from coarse to fine resolution, Instant NGP\ \cite{muller2022instant} employs multi-resolution voxel grids of trainable features that enable end-to-end training. It presents a multi-resolution hash encoding to map grids to fixed-size feature arrays from coarse to fine resolution, see Figure\ \ref{fig:fristimg2}, and the feature arrays are cached in a hash table, which represents a scene more compactly without sacrificing rendering quality. By storing the feature grids in memory, the computational burden on the MLP is reduced and training speed is significantly increased. While hash-based multi-resolution grid representation can be fast to train and therefore well suited to fast operation, trainable features at coarse resolution levels have lower resolutions and are less compressed than feature grids at fine resolution levels, resulting in limited representation power and low memory efficiency. In this paper, we aim to move toward developing a multi-resolution hybrid encoding for memory-efficient and high-quality scene representation as well as fast rendering by exploiting the different scene representation properties of positional encodings and parametric grid-based encodings. 

We propose Hyb-NeRF, a novel radiance field representation that is end-to-end optimizable for compact and fast reconstruction, and also capable of learning an accurate scene representation for novel view synthesis. Our key ideas are as follows. First, to optimize neural implicit map representations with less memory footprint, we propose coarse-to-fine hybrid features to model the scene geometry details and colors. Instead of using parametric feature grids to represent the scene at coarse levels, we present a learnable positional encoding with much fewer trainable parameters for coarse-level representation. As shown in Figure\ \ref{fig:fristimg2}, the learnable positional features are then concatenated with the fine-level multi-resolution feature grids for multi-resolution scene representation, yielding detailed 3D reconstructions and high-quality rendering. The learnable positional encoding can adaptively learn the weights of positional features and improve memory efficiency and rendering quality. Second, we embed cone tracing-based positional features in the learning of positional feature weights, which helps to significantly disambiguate the optimization process and eliminate aliasing artifacts. 

Our Hyb-NeRF can effectively reduce memory usage and enable fast high-fidelity novel view synthesis. We extensively evaluate our method with various settings and compare it with several state-of-the-art view synthesis methods in terms of the model size, rendering speed, and quality on three benchmark datasets including both synthetic and real-world scenes. All Hyb-NeRF models can reconstruct high-quality radiance fields in 9 min. Our smallest model with 8.4M trainable parameters takes 4 min to achieve better-rendering quality than state-of-the-art methods while requiring substantially less memory than previous and concurrent voxel-based methods. Our contributions are summarized as follows:

\begin{itemize}
\item We present Hyb-NeRF, a novel multi-resolution hybrid encoding that brings together the benefits of positional features and hash-based feature grids to scene representation, enabling memory-efficient, fast, and high-quality rendering. 
\item We design a learnable positional encoding that controls positional features with much fewer learnable weights to capture more geometry details at coarse resolution levels and improve rendering quality.
\item We introduce cone tracing-based features in the learning of positional feature weights, which enables our encoding to work accurately and robustly at different scales.
\end{itemize}
\section{Related Work}

\textbf{Novel View Synthesis:} The task of synthesizing images from novel viewpoints given a set of photographs has been widely studied in the field of computer graphics. Various scene representation methods have been proposed to predict an underlying geometric or image-based representation that enables rendering from novel viewpoints. Light field representations\ \cite{mildenhall2019local,attal2022learning,kalantari2016learning} directly synthesize novel views by filtering and interpolating the input images, but require high sampling rates and very dense scene capture. To render novel views from sparsely captured images, some approaches leverage pre-computed proxy geometry of the scene\ \cite{goesele2010ambient,penner2017soft}. Mesh-based methods\ \cite{newcombe2010live,waechter2014let} represent the scene with surfaces and allow rendering in real-time. However, it is difficult to optimize a mesh to capture fine geometry and topological information. Volumetric representations, such as voxel grids and multi-plane images, are better suited for gradient-based optimization and can synthesize higher-quality views than mesh-based methods. Recently, convolutional neural networks have been employed to estimate voxel grids\ \cite{rematas2020neural, sitzmann2019deepvoxels, he2020deepvoxels++, lombardi2019neural} and point clouds \ \cite{le2020novel,song2020deep,wiles2020synsin} for inward-facing captures and multi-plane images\ \cite{flynn2019deepview, zhou2018stereo, choi2019extreme} for forward-facing captures. These discrete representations are effective for view synthesis but do not scale well to higher-resolution imagery. In contrast, recent neural representations for novel view synthesis do not suffer from discretionary artifacts as they encode the scene geometry and appearance as a continuous volume\ \cite{mildenhall2021nerf,sitzmann2019scene}.

\textbf{Neural Radiance Fields:} NeRF\ \cite{mildenhall2021nerf} maps the geometry and appearance of the scene into the weights of MLPs. To assist MLPs in capturing high-frequency variations in geometry and appearance and infer high-quality novel views, NeRF encodes the input coordinate to a higher-dimensional space of multi-resolution Fourier features using positional encoding\ \cite{tancik2020fourier}. Subsequent efforts have extended NeRF for various applications, e.g., relighting\ \cite{srinivasan2021nerv,verbin2022ref,boss2021nerd}, large-scale scene modeling\ \cite{barron2022mip,zhang2020nerf++,reiser2023merf}, dynamic scene modeling\ \cite{pumarola2021d,fridovich2023k,park2021nerfies}, and deformation\ \cite{yuan2022nerf,pengcagenerf}. However, NeRF has limited reconstruction quality with sampling and aliasing issues. Many recent works have proposed to address these issues\ \cite{isaac2022exact,barron2022mip,barron2021mip}. Our method is more closely related to Mip-NeRF\ \cite{barron2021mip} that casts cones instead of rays to consider the shape and size of volume viewed by each ray and represents the volume covered by each conical frustum using an integrated positional encoding. While NeRFs are effective for photo-realistic view synthesis, they are often computationally expensive and impractical for fast rendering. In this work, we introduce a learnable positional encoding that assists our model to achieve accurate scene representation and fast rendering with shallow MLPs.

\textbf{NeRFs with explicit volumetric representations:} Recent approaches combine NeRFs with explicit volumetric representations, such as octree\ \cite{liu2020neural}, voxel grids\ \cite{sun2022direct}, Tri-Plane\ \cite{fridovich2023k,chan2022efficient} and factorization tensors\ \cite{chen2022tensorf,chen2023factor}, to reduce the size of MLPs and thus the time of training and inference. These approaches store trainable parameters in grids and interpolate these parameters to produce a continuous representation of the scene. Although these approaches use many more parameters than implicit representations, their MLPs are much smaller and can be trained to converge much faster. To represent scenes with a simple grid-based model at high resolution without prohibitive memory requirements, a series of works adopt a multistage training strategy to learn a sparse data structure\ \cite{fridovich2022plenoxels,yu2021plenoctrees} from coarse to fine. For instance, NSVF\ \cite{liu2020neural} learned a sparse voxel structure progressively to encode local properties and achieve efficient and high-quality rendering. DVGO\ \cite{sun2022direct} first learned to find a coarse geometry and then a post-activated density voxel grid was used in the second stage for generating fine details. TensoRF\ \cite{chen2022tensorf} decomposed a 4D tensor into low-rank components and applied coarse-to-fine reconstruction to achieve high memory compactness and fast rendering. Unlike updating a data structure periodically, Instant NGP\ \cite{muller2022instant} proposed a multi-resolution hash encoding method that stores feature vectors of multi-resolution grids in a compact hash table and enables one-stage end-to-end training with shallow MLPs. While this multi-resolution dense grid-based representation increases rendering speed drastically, it still requires a large memory footprint due to a large number of parameters used in both low and high-resolution grids. In contrast, we design a multi-resolution hybrid encoding that replaces feature grids at coarse resolution levels with parametric positional features and enables one-stage end-to-end training, resulting in more compact modeling and faster reconstruction while achieving even higher-quality rendering.

\section{Preliminaries}

This section provides the relevant background on neural radiance fields (NeRF)--based volume rendering using positional encodings and a multi-resolution hash encoding.

To represent a 3D scene with implicit fields for novel view synthesis, existing NeRFs map a 3D position $\mathbf{x}$ and a 2D viewing direction $\mathbf{d}$ to the corresponding density $\sigma$ and 3D color value $\mathbf{c}$ using two MLPs $\mathcal{F}_{\mathbf{w}_\theta}$ and $\mathcal{F}_{\mathbf{w}_\phi}$:
\begin{equation}
\mathcal{F}_{\mathbf{w}_\theta}: \mathbf{x}\rightarrow (\sigma, \mathbf{e}),
\end{equation}
\begin{equation}
\mathcal{F}_{\mathbf{w}_\phi}: (\mathbf{e}, \mathbf{d})\rightarrow \mathbf{c},
\end{equation}
where $\mathbf{w}_\theta$ and $\mathbf{w}_\phi$ are the weights of $\mathcal{F}_{\mathbf{w}_\theta}$ and $\mathcal{F}_{\mathbf{w}_\phi}$, respectively, and $\mathbf{e}$ is an intermediate embedding to help the MLP $\mathcal{F}_{\mathbf{w}_\phi}$ to predict color $ \mathbf{c}$. To render an image, the predicted color of a pixel $\hat{C}(\mathbf{r})$ is computed by casting a ray $\mathbf{r}=\mathbf{o}+t\mathbf{d}$ (where $\mathbf{o}$ denotes the camera origin and $t$ is the distance from the origin along the ray) into the volume and accumulating the color over $N$ point samples taken along the ray\ \cite{mildenhall2021nerf}:
\begin{equation}
\hat{C}(\mathbf{r})=\sum_{i=1}^{N}T_{i}(1-\exp(-\sigma_{i}\delta_{i}))\mathbf{c}_{i},
\label{ren}
\end{equation}
where
\begin{equation}
T_{i}=\exp\left(-\sum_{j=1}^{i-1}\sigma_{j}\delta_{j}\right).
\end{equation}
Where $\delta_{i}$ is the distance between the $i$-th pair of adjacent samples. To enable the MLPs to capture the high-frequency details from the low-dimensional inputs, i.e., $\mathbf{x}$ and $\mathbf{d}$, the inputs are projected into a higher-dimensional space by encoding functions.

\textbf{Encodings:} Given a position $\mathbf{x}$ and a viewing direction $\mathbf{d}$, NeRF\ \cite{mildenhall2021nerf} first transforms each from $\mathbb{R}$ into a higher-dimensional space $\mathbb{R}^{2L}$ using positional encoding-based Fourier features\ \cite{mildenhall2021nerf,tancik2020fourier}. Instead of casting a single infinitesimally narrow ray per point, integrated positional encoding\ \cite{barron2021mip} casts a cone from each point and controls the decay of the high-frequency Fourier features by approximating the conical frustums as a Gaussian distribution and embedding them into the positional encoding. The choice of the 3D conical frustum can significantly reduce aliasing artifacts. The fixed positional encodings are subsequently consumed by two large MLPs to estimate color and density. To reduce the size of the MLPs and save rendering time, hash encoding\ \cite{muller2022instant} maps a cascade of grids to features through a spatial hash function\ \cite{muller2022instant} and trilinear interpolation. Since the features are stored as trainable parameters, the size of the MLPs can be significantly reduced and both the training and rendering times can be saved when compared to functional encoding-based representations \ \cite{mildenhall2021nerf, barron2021mip}. 

\section{Method}
\begin{figure*}[h]
\begin{center}
\includegraphics[width=0.9\linewidth]{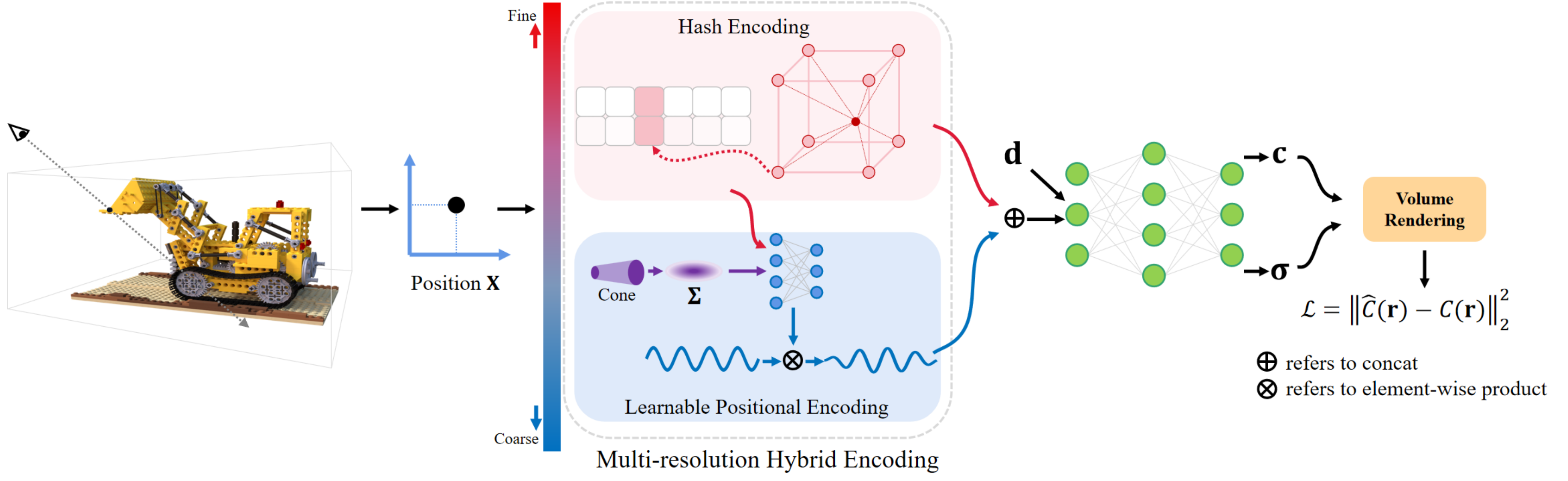}
\end{center}
\caption{Illustration of our Hyb-NeRF. Given an input 3D position $\mathbf{x}$, we encode $\mathbf{x}$ to hybrid features from coarse-to-fine resolution levels. For coarse levels, we design a learnable positional encoding to map the position $\mathbf{x}$ into parametric Fourier features. For fine levels, we map $\mathbf{x}$ to parametric features using a multi-resolution hash encoding function. The concatenated feature vectors from all levels are used to predict color and density by two shallow MLPs.}
\label{fig:posenc}
\end{figure*}

In this section, we describe Hyb-NeRF in detail. Given a set of scenes with a collection of images and their camera parameters, Hyb-NeRF learns a neural rendering model that enables photo-realistic novel view synthesis. Hyb-NeRF encodes the position $\mathbf{x}$ from $L$ multiple resolution levels with trainable encoding parameters. At the coarse levels, a learnable positional encoding is designed to adaptively map the position $\mathbf{x}$ from low-dimension space $\mathbb{R}$ to a higher-dimension space $\mathbb{R}^{2L}$, resulting an embedding vector $\gamma^{coarse}(\mathbf{x}; \mathbf{\alpha})$ with trainable parameters $\mathbf{\alpha}$. At the fine levels, we model the high-frequency details with multi-resolution parametric feature grids $\gamma^{fine}(\mathbf{x};\mathbf{\vartheta})$ with trainable parameters $\mathbf{\vartheta}$. The multi-resolution hybrid encoding of the position $\mathbf{x}$ is the concatenation of the coarse-level encoding and the fine-level encoding, i.e., $\gamma^{hyb}(\mathbf{x}; \mathbf{\alpha}, \mathbf{\vartheta})=[\gamma^{coarse}(\mathbf{x}; \mathbf{\alpha}), \gamma^{fine}(\mathbf{x};\mathbf{\vartheta})]$. The direction $\mathbf{d}$ is transformed into the spherical harmonic features $\xi(\mathbf{d})$ using the spherical harmonic basis. The encoding results of the position $\mathbf{x}$ and direction $\mathbf{d}$ are then fed into two concatenated shallow MLPs $\mathcal{F}_{\mathbf{w}_\theta}$ and $\mathcal{F}_{\mathbf{w}_\phi}$ to produce implicit fields. Later, the implicit fields with the estimated densities and colors are used for volume rendering. The overview of our Hyb-NeRF is illustrated in Figure\ \ref{fig:posenc}.

\textbf{Fine-level Encoding:} The goal of fine-level encoding is to capture the high-frequency geometric details in a scene. To realize it, we adopt multi-resolution hash-based feature grids $\gamma^{fine}(\mathbf{x};\mathbf{\vartheta})=\left\{\gamma_{l}^{fine}(\mathbf{x};\vartheta_{l})\right\}_{l=1}^{L_{f}}$. The spatial resolution of each level $\gamma_{l}^{fine}(\mathbf{x};\vartheta_{l})$ is set between the coarsest $N_{min}$ and the finest resolution $N_{max}$:
\begin{equation}
\label{res_m}
N_{l} :=\lfloor N_{min}b^{l}\rfloor,
\end{equation}
where
\begin{equation}
b :=\exp\left(\frac{\ln N_{max}-\ln N_{min}}{L_f-1}\right).
\end{equation}

\textbf{Coarse-level Encoding:}  At coarse resolution levels, the encoding function aims to model the coarse scene geometry and scene layout. While the high-frequency geometric details can be obtained by the multi-resolution hash encoding at fine resolution levels, coarse-level feature grids with trainable parameters have limited representation power. The positional encoding can map the low-dimensional position $\mathbf{x}$ into a sparser, higher-dimensional space without any trainable parameters. Combined with a large MLP, positional encodings achieve high-quality scene representation with slow training and rendering speed. To bring together the benefits of both feature grids and positional encodings to the fast high-quality novel view synthesis, we propose a learnable positional encoding for coarse-level scene representation. Specifically, we transform the input position $\mathbf{x}$ into sinusoidal activations across $L_c$ different frequency levels using a fixed positional encoding\ \cite{mildenhall2021nerf, tancik2020fourier}:

\begin{equation}
\gamma_{p}(\mathbf{x})=\left[\sin(\mathbf{x}), \cos(\mathbf{x}), \cdots, \sin(2^{L_{c}-1}\mathbf{x}), \cos(2^{L_{c}-1}\mathbf{x})\right].
\label{pe}
\end{equation}
$\gamma_{p}(\mathbf{x})$ is a continuous, multi-scale, periodic representation of $\mathbf{x}$ along each coordinate. To smoothly represent the scene from coarse-to-fine levels, the setting of the level of the positional encoding $L_c$ is based on the coarsest resolution $N_{min}$ as:
\begin{equation}
N_{min}\leq2^{L_c}.
\end{equation}

Since the frequency distribution of the local signal may vary with coordinates and shallow MLPs will cause a performance bottleneck of the fixed positional encoding, we introduce learnable weights $\mathbf{\alpha}$ to adaptively control the Fourier features to adapt the variation and improve the representation power. We use a single-layer MLP $\mathcal{F}_{\mathbf{w}_{p}}$ with trainable parameters $\mathbf{w}_{p}$ and a tanh active function to learn the weights of the Fourier features. Inspired by recent work of embedding cones into the fixed positional encoding in integrated positional encoding (IPE)\ \cite{barron2021mip}, we embed the cone tracing-based features into the learning of weights $\mathbf{\alpha}$ that assists the coarse-level encoding in representing texture with less aliasing artifacts. In addition, to adaptively capture local details according to the position, we also embed the fine-level multi-resolution feature grids into the learning of the weights ${\alpha}$. To this end, the MLP $\mathcal{F}_{\mathbf{w}_{p}}$ takes the concatenation of the multi-resolution hash-based feature grids $\gamma^{fine}(\mathbf{x};\mathbf{\vartheta})$ and the cone tracing-based Fourier features $\gamma_{p}(f(\mathbf{x}))$ as input. The weights $\mathbf{\alpha}$ can be learned as:

\begin{equation}
\mathcal{F}_{\mathbf{w}_p}: (\gamma^{fine}(\mathbf{x};\mathbf{\vartheta}),\gamma_{p}(f(\mathbf{x})))\rightarrow \mathbf{\alpha},
\end{equation}
where $f(\mathbf{x})$ is the cone tracing-based transformation of the point $\mathbf{x}$. We adopt cone tracing-based features by casting a cone from each pixel and approximating a conical frustum with a multivariate Gaussian\ \cite{barron2021mip}. The covariance of the final multivariate Gaussian $\mathbf{\Sigma}$ is given by
\begin{equation}
\mathbf{\Sigma}=\sigma_{t}^{2}(\mathbf{d}\mathbf{d}^{T})+\sigma_{r}^{2}\left(\mathbf{I}-\frac{\mathbf{d}\mathbf{d}^{T}}{\left\|\mathbf{d}\right\|_{2}^{2}}\right),
\end{equation}
where $\sigma_{t}^{2}$ and $\sigma_{r}^{2}$ are the variances along the ray and the perpendicular direction of the ray, respectively. Instead of computing the diagonal of the covariance matrix and integrating only the axis-aligned Fourier features within the Gaussian cone, here we compute the full upper triangular elements of the covariance matrix:
\begin{equation}
 f(\mathbf{x})=\text{triu}(\mathbf{\Sigma}).
\end{equation}
This remains the non-axis-aligned/non-diagonal components and improves the representation quality of the MLPs. We map the input $f(\mathbf{x})$ into a higher-dimensional space using the fixed positional encoding in Eq.(\ref{pe}) and obtain our final cone tracing-based Fourier features $\gamma_{p}(f(\mathbf{x}))$. 
The coarse-level encoding can be obtained by weighting the fixed positional encoding as:
\begin{equation}
\gamma^{coarse}(\mathbf{x}, \mathbf{\alpha})=\gamma_{p}(\mathbf{x})\otimes \mathbf{\alpha}
\end{equation}
where $\otimes$ is element-wise multiplication. 

\textbf{Model Architecture:}
We employ two concatenated MLPs $\mathcal{F}_{\mathbf{w}_{\theta}}$ and $\mathcal{F}_{\mathbf{w}_{\phi}}$ to map each encoded position and view direction into its corresponding volume density $\sigma$ and color $\mathbf{c}$:
\begin{equation}
\mathcal{F}_{\mathbf{w}_\theta}: \gamma^{hyb}(\mathbf{x};\mathbf{\alpha}, \mathbf{\vartheta})\rightarrow (\sigma, \mathbf{e}),
\end{equation}
\begin{equation}
\mathcal{F}_{\mathbf{w}_\phi}: (\mathbf{e}, \xi(\mathbf{d}))\rightarrow \mathbf{c}.
\end{equation}
The estimated density and color are then used to predict pixel color $\hat{C}(\mathbf{r})$ as in (\ref{ren}). Our model is optimized end-to-end by minimizing the $L_{2}$ reconstruction loss between the rendered pixel color $\hat{C}(\mathbf{r})$ and the ground truth color $C(\mathbf{r})$:
\begin{equation}
\mathcal{L}=\left\|C(\mathbf{r})-\hat{C}(\mathbf{r})\right\|_{2}^{2}.
\end{equation}
\section{Experiments}
\begin{table*}[h]
\begin{center}
\resizebox{0.9\textwidth}{0.175\linewidth}{
\begin{tabular}{lccccccccccc}
\toprule
\multirow{2}{*}{\textbf{Method}}&\multicolumn{6}{c}{\textbf{Blender}}&\multicolumn{2}{c}{\textbf{Synthetic-NSVF}}&\multicolumn{2}{c}{\textbf{Tanks\&Temples}} \\
\cmidrule(r){2-7} \cmidrule(r){8-9} \cmidrule(r){10-11}
&\textbf{\#Params$\downarrow$} &\textbf{Time$\downarrow$} &\textbf{Iters$\downarrow$} &\textbf{PSNR(dB)$\uparrow$} & \textbf{SSIM$\uparrow$} &\textbf{LIPIS$\downarrow$} &\textbf{PSNR(dB)$\uparrow$} & \textbf{SSIM$\uparrow$} &\textbf{PSNR(dB)$\uparrow$} & \textbf{SSIM$\uparrow$}\\
\midrule
NeRF \ \cite{mildenhall2021nerf}&\colorbox{orange!40}{1191k} &3h &- &31.01 & 0.947& 0.081&29.97 & 0.944&25.78 & 0.864 \\
NSVF \ \cite{liu2020neural}&3-16M &\textgreater48h &- &31.75 & 0.953& \colorbox{orange!40}{0.047}&34.47 & 0.976&28.48 & 0.901 \\
MipNeRF \ \cite{barron2021mip}&\colorbox{red!40}{612K} &2.8h &- &33.09 & 0.947& \colorbox{red!40}{0.043} &- &- & - &- & - \\
DVGO \ \cite{sun2022direct}&$\textgreater$25M &15m &\colorbox{orange!40}{30k} &31.95 & 0.957& 0.053 &34.51 & 0.972&28.41 & 0.911 \\
TensoRF \ \cite{chen2022tensorf}&17M &17m &\colorbox{orange!40}{30k} &33.14 & \colorbox{orange!40}{0.963}& \colorbox{orange!40}{0.047}&\colorbox{yellow!40}{36.24} & 0.981&28.56 & \colorbox{yellow!40}{0.920} \\
Instant NGP \ \cite{muller2022instant}&12.6M &\colorbox{red!40}{4min} &\colorbox{orange!40}{30k} &32.59 & 0.960& 0.053&- & -&- & - \\
\hline
\multirow{2}{*}{JNGP\ \cite{yang2023jnerf}} &12.6M &\colorbox{orange!40}{5min} &\colorbox{yellow!40}{40k} &32.67 & 0.959& 0.054&34.91 & 0.976&27.95 &0.916 \\
&24.4M &\colorbox{yellow!40}{7.5min} &\colorbox{yellow!40}{40k} &32.96 & \colorbox{orange!40}{0.963}& 0.051 &35.71 & \colorbox{yellow!40}{0.983}&28.11 & \colorbox{orange!40}{0.921} \\
\hline
Hyb-NeRF &\colorbox{yellow!40}{8.4M} &\colorbox{red!40}{4min} &\colorbox{red!40}{21k} &33.28 & 0.960& 0.055&35.68 & 0.981&28.34 & 0.909 \\
(early-stop)&16.8M &\colorbox{orange!40}{5min} &\colorbox{red!40}{21k} &\colorbox{orange!40}{33.79} & \colorbox{red!40}{0.964}& \colorbox{yellow!40}{0.049}&\colorbox{orange!40}{36.72} & \colorbox{orange!40}{0.984}&\colorbox{yellow!40}{28.58} & 0.915 \\
\hline
Hyb-NeRF&\colorbox{yellow!40}{8.4M} &\colorbox{yellow!40}{7.5min} &\colorbox{yellow!40}{40k} &\colorbox{yellow!40}{33.49} & \colorbox{yellow!40}{0.961}& 0.053&36.27 & 0.982&\colorbox{orange!40}{28.70} & 0.915 \\
(fully-trained)&16.8M &9min &\colorbox{yellow!40}{40k} &\colorbox{red!40}{33.94} & \colorbox{red!40}{0.964}& \colorbox{orange!40}{0.047}&\colorbox{red!40}{37.14} & \colorbox{red!40}{0.985}&\colorbox{red!40}{29.04} & \colorbox{red!40}{0.922}\\
\bottomrule
\end{tabular}}
\end{center}
\caption{Quantitative comparison on Blender, Synthetic-NSVF, and Tanks\&Temples. We also report comparison results of the average training times, the amounts of parameters, and iteration steps for the Blender dataset. Our method achieves the best rendering quality with efficient memory use.}
\label{table:Result}
\end{table*}

This section evaluates the proposed Hyb-NeRF on rendering of synthetic and realistic scenes. We compare our method with previous state-of-the-art methods quantitatively and qualitatively and provide extensive ablation studies to validate different options in encoding designs. In addition, we also provide the rendering performance of our model with different training times and amounts of parameters. 

\textbf{Datasets:} We conduct our experiments on three datasets including Blender\ \cite{mildenhall2021nerf}, Synthetic-NSVF\ \cite{liu2020neural} and Tanks\&Temples\ \cite{liu2020neural,Knapitsch2017}. 1) Blender: It consists of 8 synthetic scenes and each scene has an object (\emph{chair, drums, ficus, hotdog, lego, materials, mic, and ship}) with 400 synthesized images and their corresponding camera parameters. 2) Synthetic-NSVF: Similar to Blender, Synthetic-NSVF contains synthetic scenes with more complex physical structures. We use a subset of five scenes (\emph{bike, palace, robot, toad, and wineholder}) in our experiments. Each scene has a set of synthesized images of an object and camera poses. For both Blender and Synthetic-NSVF, the image resolution is ${800}\times{800}$ pixels, and we follow the setups in NeRF and NSVF to use 100 views for training and 200 for testing. 3) Tanks\&Temples: It is a benchmark real-world dataset for image-based 3D reconstruction. We use a subset of the provided scenes (\emph{Ignatius, Truck, Barn, Caterpillar, and Family}), each containing views captured by an inward-facing camera circling. We follow the default split to produce training and testing views and the resolution of each view is ${1920}\times{1080}$ pixels.

\textbf{Baselines:} We compare our method to the following baseline methods: NeRF\ \cite{mildenhall2021nerf}, NSVF\ \cite{liu2020neural}, Mip-NeRF\ \cite{barron2021mip}, DVGO\ \cite{sun2022direct}, TensoRF\ \cite{chen2022tensorf}, Instant NGP\ \cite{muller2022instant}. Since we implement our Hyb-NeRF on top of a Jittor\ \cite{hu2020jittor} reimplementation of Instant NGP, we also include this implementation as our baseline and refer to it as JNGP\ \cite{yang2023jnerf}.

\textbf{Implementation Details:} We use $L_c=8$ frequency bands for our learnable positional encoding $\gamma^{coarse}(\mathbf{x})$ at coarse levels. For the hash encoding performed at high resolutions, the setting of the hash table length $T$ and the dimension of the feature vectors $F$ at each level are the same as Instant NGP ($T=2^{19}$ and $F=2$). The density MLP $\mathcal{F}_{\mathbf{w}_\theta}$ is a single-hidden-layer network and the color MLP $\mathcal{F}_{\mathbf{w}_\phi}$ is a two-hidden-layer network. For both, each layer contains 64 channels. We set the coarsest resolution $N_{min}=180$ and use two different setups for $L_f$, i.e., $L_f=8$ and $L_f=16$, resulting in $8.4$ million (M) and $16.8$M trainable parameters of our model, respectively. We also include the rendering results of JNGP\ \cite{yang2023jnerf} with two different $L_f$, i.e., $L_f=16$ and $L_f=32$, which produce $12.6$M and $24.4$M parameters, respectively. It is valuable to mention that the coarsest resolution $N_{min}$ in Instant NGP is set to $16$, which is much smaller than ours. Since our model uses a larger $N_{min}$ and a smaller $L_f$, it contains much fewer trainable parameters than Instant NGP. We use a batch size of $256$K samples and the Adam optimizer\ \cite{kingma2014adam} with $\beta_{1}=0.9$ and $\beta_{2}=0.99$. We train our model with a learning rate of $5\times10^{-3}$ and the same learning rate decay schedule as Instant NGP. Our experiments are run on a PC with a single NVIDIA Geforce RTX3090 GPU (24GB).

\subsection{Comparisons}
\begin{figure*}[h]
\begin{center}
\includegraphics[width=0.7\linewidth]{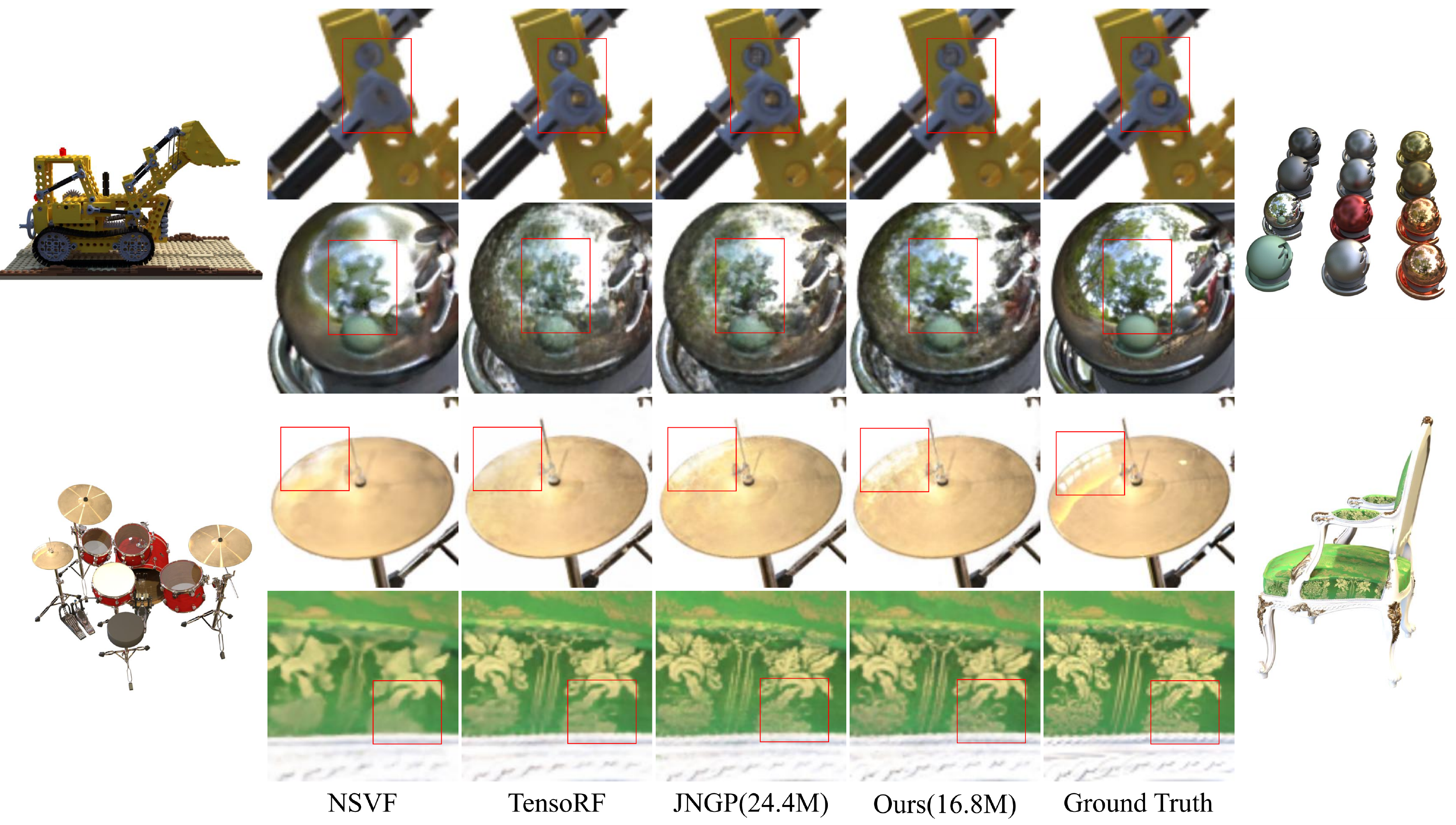}
\end{center}
\caption{Visual comparison between our fully-trained model with NSVF\ \cite{liu2020neural}, TensoRF\ \cite{chen2022tensorf} and JNGP\ \cite{yang2023jnerf} on four synthetic scenes in the Blender dataset. Our model synthesizes the most photo-realistic novel views with finer detail.}
\label{fig:showoff}
\end{figure*}
\textbf{Quantitative Comparison:} We report our quantitative comparison results in Table\ \ref{table:Result} in terms of the peak signal-to-noise ratio (PSNR) and structural similarity index (SSIM) on the three datasets. To evaluate the effect of training time and model size on rendering accuracy, we also report corresponding optimization iterations (the optimization iterations of NeRF, NSVF, and MipNeRF are not present, as this comparison would not be particularly meaningful), training time as well as the number of trainable parameters. We provide rendering results of our method with different numbers of parameters and training times, including two early-stop models and two fully-trained models. We observe that our early-stop model with 8.4M parameters can obtain comparable numerical performance to most of the baselines. Increasing the number of resolution level $L_f$ from $8$ to $16$ further improves the reconstruction accuracy of our method. Our early-stop model with 16.8M parameters performs better than other methods on synthetic datasets and performs nearly on par with the state-of-the-art neural implicit model on realistic scenes while using less training time. In addition, our model trained with 16.8M parameters in 9 minutes outperforms other methods with fewer errors on both synthetic and real-world scenes. 

\begin{figure*}[h]
\begin{center}
\includegraphics[width=0.7\linewidth]{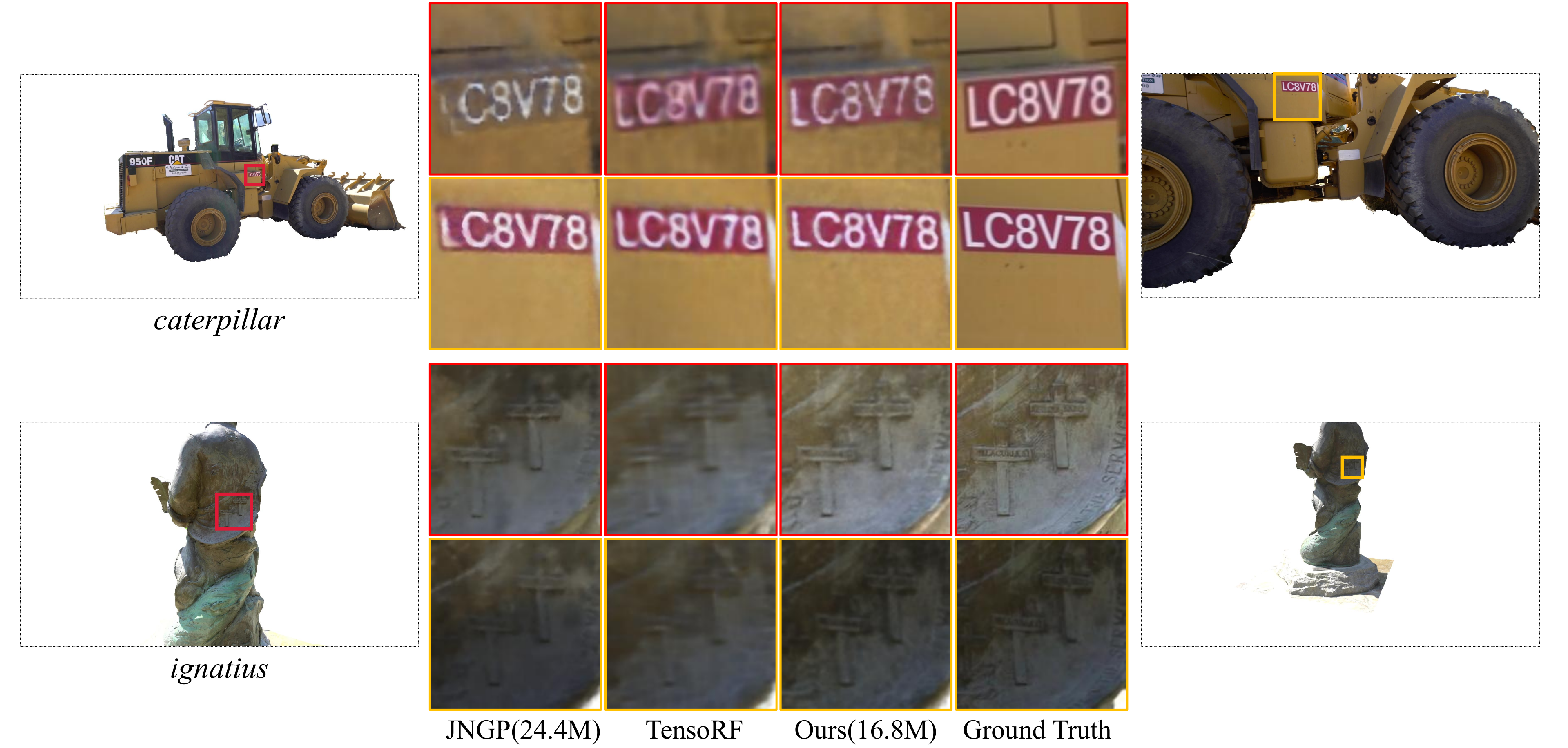}
\end{center}
\caption{Qualitative comparison results between our fully-trained model, JNGP\ \cite{yang2023jnerf} and TensoRF\ \cite{chen2022tensorf} on \emph{caterpillar} and \emph{ignatius} scenes in the Tanks\&Temples dataset. Our model reconstructs better physical details and colors at different scales.}
\label{fig:showtanks}
\end{figure*}

\textbf{Qualitative Comparison:} Figure\ \ref{fig:showoff} shows the visual comparison results on synthetic scenes in Blender. It shows that our model can recover the finer appearance and geometric details, e.g., \emph{lego}'s holes, reflections on the surface of \emph{materials} and \emph{drums}, \emph{chair}'s color and texture. The visual comparison results on real-world scenes in Tanks\&Temples are shown in Figure\ \ref{fig:showtanks}. We observe that TensoRF synthesizes reasonable views but still contains blurry textures. JNGP synthesizes views with more texture details but struggles to handle the varying exposure and inconsistent masks, resulting in chromatic aberrations. Since our model embeds conical frustums, it obtains more visually realistic colors while reconstructing surface detail well. 

\subsection{Ablation Studies} 

\textbf{Effect of the learnable positional encoding:} We perform ablation evaluations on the Blender dataset to validate the effectiveness of design choice on learning positional features and the concatenation of the learnable positional features and dense feature grids from coarse-to-fine resolution levels. First, we compare the rendering quality from the baseline JNGP model and two variants of Hyb-NeRF: one is directly concatenating the original fixed positional encoding at coarse levels (\textbf{Hyb-NeRF, fixed PE}) and another is concatenating the learnable positional encoding that only uses the fine-level hash-based feature grids for parameter learning (\textbf{Hyb-NeRF, learnable PE w hash encoding}). Quantitative results in Table\ \ref{table:ablation} show that the model with fixed PE (\textbf{Hyb-NeRF, fixed PE}) outperforms JNGP, but achieves lower rendering quality than all models with learnable positional encoding (\textbf{Hyb-NeRF, learnable PE w hash encoding}; \textbf{Hyb-NeRF, learnable PE w cone}; \textbf{Hyb-NeRF}). This is because fixed positional features at coarse levels guarantee the convergence of \textbf{Hyb-NeRF, fixed PE}, and all trainable feature grids at fine levels help to capture better geometry details. \textbf{Hyb-NeRF, learnable PE w hash encoding} achieves better results than \textbf{Hyb-NeRF, fixed PE}, since the position-related fine-resolution feature grids allow the learnable positional encoding to adaptively capture local details, enabling accurate scene representation with the shallow MLPs.

\textbf{Effect of the cone-tracing embedding:} We also evaluate the effect of the cone-tracing embedding. We compare the rendering quality from Hyb-NeRF with and without embedding the cone tracing-based features in the learning of positional feature weights (\textbf{Hyb-NeRF, learnable PE w cone}; \textbf{Hyb-NeRF, learnable PE w hash encoding}). As shown in Table\ \ref{table:ablation}, the embedding of the cone tracing-based features in $\gamma^{coarse}(\mathbf{x};\mathbf{\alpha})$ indeed eliminates aliasing artifacts and improves rendering quality. Meanwhile, we show that a direct concatenation of the multi-resolution hash encoding and IPE at all resolution levels (\textbf{JNGP, cat. IPE}) provides better rendering quality than JNGP, but lower rendering quality than our Hyb-NeRF models. This is because IPE as an additional input encoding of the multi-resolution hash encoding can assist the MLPs in capturing more geometry details. However, this performance boost is limited, since the shallow MLPs use IPE with untrainable features and have limited representation power. Our cone-tracing embedding strategy (\textbf{Hyb-NeRF, learnable PE w cone}) provides a significant rendering quality improvement over other embedding design choices. 

\begin{table}[h]
\begin{center}
\resizebox{0.44\textwidth}{0.18\linewidth}{
\begin{tabular}{lcccc}
\toprule
\textbf{Method}&\textbf{PSNR(dB)}$\uparrow$ &\textbf{SSIM$\uparrow$}\\
\midrule
JNGP & 32.67 & 0.959\\
JNGP, cat. IPE & 33.09 & 0.960 \\
\midrule
Hyb-NeRF, fixed PE &32.83 & 0.960 \\
Hyb-NeRF, learnable PE w hash encoding &\colorbox{yellow!40}{33.19} & \colorbox{yellow!40}{0.961} \\
Hyb-NeRF, learnable PE w cone &\colorbox{orange!40}{33.53} & \colorbox{orange!40}{0.962} \\
Hyb-NeRF &\colorbox{red!40}{33.94} &\colorbox{red!40}{0.964} \\
\bottomrule
\end{tabular}}
\end{center}
\caption{Quantitative ablation study results on Blender. We compare our model with variants in terms of PSNR and SSIM. All models use 16 hash encoding levels.}
\label{table:ablation}
\end{table}

\section{Conclusion}
We proposed Hyb-NeRF, a novel neural scene representation that is end-to-end optimizable for high-quality and fast rendering. Hyb-NeRF maps the input coordinate to a hybrid encoding that includes parametric positional features at coarse resolution levels and hash-based feature grids at fine resolution levels. The parametric positional features use much fewer trainable parameters for accurate coordinate representation at coarse levels, resulting in a significantly lower memory footprint and higher rendering quality. In addition, we embed the cone tracing-based features in the learning of modulating positional features, leading to better reconstruction quality and photo-realistic novel view synthesis. We show that using a learnable multi-resolution hybrid encoding with tiny MLPs as scene representation enables our Hyb-NeRF to provide favorably against the state-of-the-art in synthesizing more realistic rendering results and efficient memory use.

\section{Acknowledgement}   
This work was supported in part by the National Natural Science Foundation of China under Grants 62106095 and 62071212, Shenzhen Science and Technology Program under Grant KCXFZ20211020174802004.
{\small
\bibliographystyle{ieee_fullname}
\bibliography{PaperForReview}
}

\end{document}